\documentclass[final]{cvpr}

\usepackage{times}
\usepackage{epsfig}
\usepackage{graphicx}
\usepackage{amsmath}
\usepackage{amssymb}
\usepackage{comment}
\usepackage[ruled,vlined,linesnumbered]{algorithm2e}

\usepackage[pagebackref=true,breaklinks=true,colorlinks,bookmarks=false]{hyperref}

\def\cvprPaperID{28} 
\def\confYear{EarthVision 2021}

\DeclareMathOperator*{\argmax}{arg\,max}

\begin{document}

\title{Addressing Visual Search in Open and Closed Set Settings}

\author{Nathan Drenkow \and Philippe Burlina \and Neil Fendley \and Onyekachi Odoemene \and Jared Markowitz\\
The Johns Hopkins University Applied Physics Laboratory\\
11100 Johns Hopkins Road, Laurel, Maryland 20723\\
{\tt\small Nathan.Drenkow@jhuapl.edu}
\\
}

\maketitle

\begin{abstract}
Searching for small objects in large images is a task that is both challenging for current deep learning systems and important in numerous real-world applications, such as remote sensing and medical imaging.  Thorough scanning of very large images is computationally expensive, particularly at resolutions sufficient to capture small objects.  The smaller an object of interest, the more likely it is to be obscured by clutter or otherwise deemed insignificant.  We examine these issues in the context of two complementary problems: closed-set object detection and open-set target search.  First, we present a method for predicting pixel-level objectness from a low resolution gist image, which we then use to select regions for performing object detection locally at high resolution.  This approach has the benefit of not being fixed to a predetermined grid, thereby requiring fewer costly high-resolution glimpses than existing methods.  Second, we propose a novel strategy for open-set visual search that seeks to find all instances of a target class which may be previously unseen and is defined by a single image. We interpret both detection problems through a probabilistic, Bayesian lens, whereby the objectness maps produced by our method serve as priors in a maximum-a-posteriori approach to the detection step.  We evaluate the end-to-end performance of both the combination of our patch selection strategy with this target search approach and the combination of our patch selection strategy with standard object detection methods.  Both elements of our approach are seen to significantly outperform baseline strategies.
\end{abstract}

\section{Introduction}
 
Artificial intelligence (AI), principally via advances in deep learning (DL),
 has recently shown great success on an ever expanding number of tasks.  In problems such as image classification~\cite{krizhevsky2017imagenet,he2016deep}, object detection~\cite{Ren2015FasterRT,Redmon2015YouOL} and image segmentation~\cite{pekala2019deep,ronneberger2015u} as well as in applications  such as medical diagnostics~\cite{burlina2019assessment, pekala2019deep}, AI approaches have met or exceeded the capabilities of humans and traditional machine learning~\cite{burlina2011automatic}. Among other things, current AI/DL research has tackled issues such as open-set recognition~\cite{scheirer2012toward,geng2020recent} (one of the foci of this study), privacy~\cite{shokri2017membership}, adversarial attacks~\cite{carlini2017adversarial}, low-shot learning~\cite{ravi2016optimization,burlina2020low}, and AI bias~\cite{burlina2021addressing}.

\begin{figure}[t!]
\centering
\includegraphics[width=\linewidth]{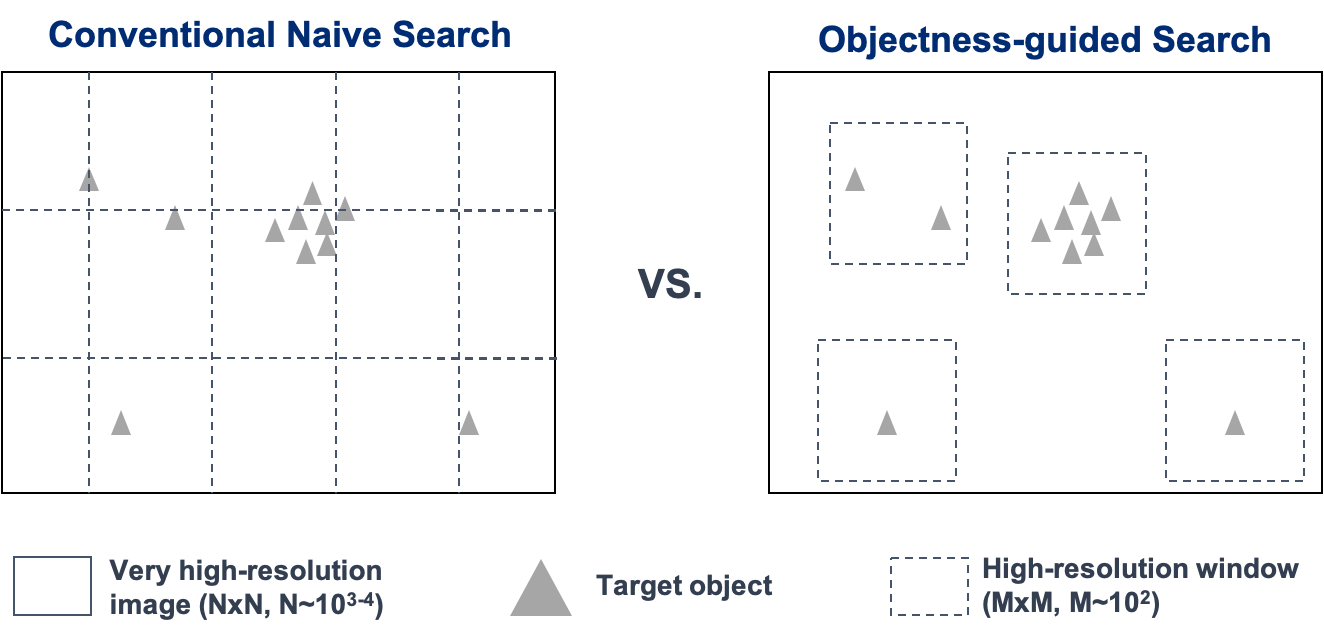}

\caption{Conventional approaches use image tiling or sliding windows to ensure coverage while keeping tile dimensions small enough to be processed by standard object detectors.  We propose to use predicted objectness to achieve the same end while minimizing the total number of high-resolution windows or glimpses required.}
\label{fig:efficiency}
\end{figure}

This study is specifically focused on using convolutional neural networks (CNNs) for object detection. Much progress has been made on this task through algorithms including YOLO and Fast(er)-RCNN~\cite{Ren2015FasterRT,Redmon2015YouOL,redmon2018yolov3,Tan_2020_CVPR}.  However, most existing object detectors rely on the assumption that the objects of interest occupy a significant fraction of the search area.  We instead consider the situation where objects may be several orders of magnitude smaller than the image size (e.g. images with thousands of pixels per side and relevant objects spanning only tens of pixels).  This occurs frequently in remote sensing applications including visual search of satellite imagery (e.g., looking for vehicles in a parking lot) and microscopy images (e.g., detecting synapses in electron microscopy imaging of brain tissue~\cite{takemura2015synaptic}). 
To date (and as discussed further in Section~\ref{sec:related_work}), most conventional DL-based object detectors have struggled when applied to remote sensing.  

To further compound the challenges associated with detecting small objects, most machine vision techniques operate on images on the order of a few hundred pixels per side, for example the $224 \times 224$ pixel images of ImageNet ~\cite{imagenet_cvpr09}.  Processing very high resolution (vHR) images requires additional computation, time, and money, particularly in streaming applications. To be useful, approaches to the detection problem must efficiently manage trade-offs among  computational cost, memory, and performance in an application-specific manner.  

Thoroughly searching such vHR images for objects appearing at varying scales motivates the development of unconventional approaches.  Standard techniques relying on CNNs involve processing sliding windows of full images.  The computational cost of this process increases quadratically with the size of the image, incurring large memory and computational footprints that eventually become prohibitive.  In practice, the computational budget may be fixed, leading to choices about how to prioritize window selection.

\subsubsection*{Objectives}
We thus pursue two aims for this work: to achieve high-performing object detection in vHR images and to develop methods that scale according to data and computational constraints.

We address these aims for two different detection scenarios: (1) \textbf{closed-set object detection}, whereby we search for instances of a fixed number of pre-identified object classes, and (2) \textbf{open-set target-guided search}, where the algorithm must find instances of a target class defined by only a single image.  Scenario (1) most closely resembles standard object detection scenarios, where a model is trained and tested on a known set of classes.  In scenario (2), the classes are not known ahead of time. During training and inference, the model is presented with a single target image from which it must infer the target class and then detect instances of that class in a vHR search image.  In this scenario, the target classes are defined by their target images at each iteration and thus do not need to be constrained to a fixed number of classes (hence the \emph{open set} nature of the problem).

To address the aforementioned challenges with object detection in vHR images for these two scenarios, we develop an approach to more efficiently ``look'' for objects belonging to classes of interest (Figure~\ref{fig:efficiency}).  Recent work ~\cite{Uzkent2020EfficientOD} has tackled aspects of this problem using deep reinforcement learning (DRL). We instead address it through an approach that identifies regions of high potential \textit{objectness}. We use a probabilistic interpretation of object detection in vHR images whereby predicted objectness maps act as priors for detection algorithms, allowing us to perform search in a maximum-a-posteriori (MAP) setting.   

\section{Related Work}
\label{sec:related_work}

Recent research has tackled the problem of object detection in visual data through the use of convolutional neural networks.  R-CNN ~\cite{girshick_rich_2014} and Fast-RCNN ~\cite{fastrcnn} rely on selective search to identify proposal regions, while Faster-RCNN ~\cite{Ren2015FasterRT} jointly identifies proposal regions and their classes.  The YOLO family of algorithms ~\cite{Redmon2015YouOL,redmon2018yolov3,bochkovskiy_yolov4_2020}, on the other hand, pass the entire image to a detection network. They produce bounding boxes and object probabilities in a single pass, leading to superior speed.  More recently, EfficientDet ~\cite{Tan_2020_CVPR} used several algorithmic innovations to provide state-of-the-art detection performance in a quantifiably more efficient manner.  However, like prior methods, EfficientDet struggles to detect small objects in large, cluttered scenes.  While all of these methods perform well on natural imagery, they are not immediately applicable to vHR images. vHR images would need to be heavily downsampled or at a minimum tiled just to enable processing through standard architectures and on conventional GPU hardware.

Early work in applying DL-based object detectors to overhead imagery~\cite{sakla2017, vanetten2018, vanetten2019, Shermeyer_2019_CVPR_Workshops, rainey2012, Liao_2020, tang2017vehicle} has focused on challenges including very large changes in scale, the need for rotation invariance, and limited amounts of training data.  While progress has been made in addressing these challenges, the issue of image resolution has largely been left to naive windowing/tiling or multi-scale approaches.  These methods show improved detection results on large images but do not address efficiency concerns.

Recent work ~\cite{Uzkent2020EfficientOD} has allowed for targeted object detection in vHR images using DRL.  This approach uses a two-stage selection process on fixed grids of potential search regions as a way to address efficiency challenges.  Each high-resolution (HR) grid tile is either downsampled or processed natively by a conventional detection network, with a learned policy being used to make the low- vs. high-resolution determination. The DRL agent is trained to choose which regions of an image to process at low resolution (LR) and which regions to process at high resolution (HR) in order to best balance efficiency with detection performance. 

Here we provide an approach that allows flexible sampling of HR windows (also referred to as \textit{glimpses}) as an alternative to a fixed-grid approach. Our method does not preclude the use of DRL to sample these glimpses, but our focus is on finding a representation of vHR images that eases glimpse selection and subsequent object detection.  To achieve this, we develop a method for estimating objectness from low-resolution (LR) \textit{gist} images that then guides our glimpse-sampling approach. Finally, we demonstrate how this approach can be used for open-set search in a MAP framework.

\section{Methods}
We discuss complementary approaches to object detection in vHR overhead images.  In particular, we address object detection in both closed and open-set scenarios, using estimates of objectness based on low-resolution gist images to guide the search and detection processes.  

\subsection{Efficient Closed-Set Detection}
\subsubsection{Objectness Estimation}
We first develop an approach that allows our model to examine the full scene at low resolution (and with low computation cost), producing a saliency map to guide subsequent high-resolution object detection. Specifically, we train a deep neural network (DNN) to produce a density map of \textit{objectness}; that is, to produce a map that encodes the likelihood of regions containing objects from the classes we wish to detect.  In practice, given a high-resolution ground-truth class-level semantic segmentation, $S_c$, we generate a binary segmentation $S_b$ by converting the class-level labels to binary labels that represent the presence or absence of an object (e.g., for each pixel, $S_b(i,j) = 1$ if $S_c(i,j) \in \{1, \dots, C\}$ and $S_b(i,j)=0$ otherwise).

To train the prediction model, we generate low-resolution gist images $I_{g}$ from the original vHR image $I_{\text{vHR}}$ by resizing from the native resolution (e.g., 4000x4000) to low-resolution (e.g., 128x128) and similarly scale $S_b$ to get a gist mask $S_g$ of fixed dimensions, $(w_{\text{gist}}, h_{\text{gist}})$.  For convenience and following common practice, we assume images are square (i.e., $w_{\text{gist}} = h_{\text{gist}} = d_{\text{gist}}$). The scale factor, $\alpha$, is defined as 

\begin{equation}
    \alpha = \frac{d_{\text{gist}}}{d_{\text{vHR}}}.
\end{equation}

\noindent As a result of the downsampling, we expect the gist images to lose some class-level information and potentially even reduce small objects beyond recognition (e.g., to one pixel). However our intent is to produce density maps that capture object likelihood, rather than a true segmentation. We find that the gist images preserve sufficient context to support those predictions (e.g., textures, co-occurring larger objects, etc.).

Using the gist images, we train a U-Net~\cite{ronneberger2015u} to produce objectness density maps $\pi$ from the gist images. Denoting the U-Net as $f$, then $f:I_g \rightarrow \pi$ where $\pi_{ij} \in [0, 1]~\forall~i,j $. We define our training loss to be the binary cross-entropy between the true object mask and the prediction,

\begin{equation}
    L_{\pi} = - \sum_{i,j\in S_g} y_{ij}\log(\pi_{ij}) - (1-y_{ij})\log(1-\pi_{ij}).
\label{eq:obj_loss}
\end{equation}

\noindent Here $y_{ij}$ is the true objectness value at pixel $(i,j)$ and $\pi_{ij}$ is the objectness prediction at that same location representing the likelihood that the corresponding location is occupied by an object of interest.

In the event that the ground-truth segmentation is unavailable, an alternative is to convert bounding box annotations to Gaussian densities of the same shape and normalized to have a peak of 1 at the center. Supplemental experiments demonstrated this to be an effective alternative when high-quality ground truth is sparse.

\subsubsection{Region Selection}
Given the objectness maps as a prior, we aim to efficiently achieve effective object detection through as few high-resolution glimpses as possible. To disentangle the benefits of the objectness map for detection from the results of using more sophisticated glimpse-selection strategies (e.g. DRL), we adopt the following simple yet highly-effective approach.
First, we define the appropriate glimpse image dimensions as determined by the downstream object detector.  Assuming square input images as before, we use $d_{\text{glimpse}}$ to denote the final glimpse resolution. From this definition, we can determine the corresponding size of a glimpse in the gist image via $d'_{\text{glimpse}} = \text{ceil}(\alpha \cdot d_{\text{glimpse}})$.

Given the glimpse size relative to the gist image, we seek to define a policy for sampling glimpses that maximizes the detection of objects of interest. We initially focus on rule-based policies that are deterministic and interpretable.  Our simple yet effective policy is described in Algorithm~\ref{alg:glimpse_policy}.  The method iteratively samples glimpses that maximize the total available objectness. This strategy allows for glimpses to overlap each other, but can be easily modified by increasing or decreasing the $\beta$ term (for instance, $\beta = -(d'^2_{\text{glimpse}})$ prevents any glimpse overlap).

At each step in the loop in Algorithm~\ref{alg:glimpse_policy} we search for where to glimpse next so that the sampled glimpse maximizes the total objectness contained within it. To speed up this search, we first compute an integral image of the current map $\pi$: 
\begin{equation}
    S_\pi(i,j) = \sum_{i'\le i, j' \le j} \pi(i',j')~~~\forall~i,j
    \label{eq:sat}
\end{equation}
 
\noindent From this table, we can easily search for the glimpse with maximum objectness by simply summing over shifted (according to $d'_{LR}$) versions of $S_{\pi}$ and finding the location of the maximum. Using Python-style notation to illustrate indexing $S_{\pi}$ for the four shift operations, the glimpse with maximum objectness over all search locations $(x,y)$ can be found according to:

\begin{equation}
\begin{split}
    p_{x,y} = \argmax_{x,y} &(S_{\pi}[:-d'_{LR}, :-d'_{LR}] + S_{\pi}[d'_{LR}:, d'_{LR}:]) ~ - \\
    &(S_{\pi}[:-d'_{LR}, d'_{LR}:] + S_{\pi}[d'_{LR}:, :-d'_{LR}])
\label{eq:max_obj}
\end{split}
\end{equation}

\subsubsection{Object Detection}
Once the LR glimpse region has been selected, the corresponding HR region from the vHR image is selected and input to a DL-based object detector (e.g., YOLO) which produces the list of detections. The object detector is pretrained to detect a closed set of object classes given tiles of the same size as the HR glimpses.  This glimpse/detection process is repeated until one of three conditions is met: (1) the entire image has been processed, (2) the cumulative objectness covered by the glimpses exceeds a pre-defined threshold (e.g., $>95\%$ objectness covered), or (3) the number of glimpses processed fulfills a pre-defined computational budget (e.g., a maximum of 10 glimpses).

\begin{algorithm}[t]
    \SetAlgoLined
    \KwResult{Selection of a set of glimpses}
    \SetKwInOut{Input}{Input}
    \SetKwInOut{Output}{Output}
    \Input{$\pi$ : Objectness map prior of dimension $d_{\text{gist}}$ \\
    $d_{\text{glimpse}}$ : Glimpse dimension in HR image-space \\
    $\alpha$ : Gist scaling factor\\
    $n_{\text{glimpse}}$ : Maximum number of glimpses \\
    $\beta$ : Objectness penalty term [default=0]}
    \Output{$\mathcal{G}$ : Set of selected glimpse locations}
    
    Compute $d'_{\text{glimpse}} = \text{ceil}(\alpha \cdot d_{\text{glimpse}})$\; 
    Compute $\sigma =\frac{d'_{\text{glimpse}}}{4}$\;
    Generate Gaussian kernel, $k$ with width $\sigma$ and dimension $d'_{\text{glimpse}}$\;
    Compute $\pi'= \text{conv}(\pi, k)$\; 
    $\mathcal{G} = \{\}$\;
    \For{$i=1:n_{\text{glimpse}}$}{
        Generate $S_{\pi'}$\ according to~(\ref{eq:sat})\;
        Compute $p_{x,y}$ according to~(\ref{eq:max_obj})\;
        Add $p_{x,y}$ to $\mathcal{G}$\;
        Reset objectness at glimpse location $\pi'[p_x:p_x + d'_{LR}, p_y:p_y + d'_{LR}] = \beta$\;
    }
    return $\mathcal{G}$    
    \caption{Glimpse Selection Policy}
    \label{alg:glimpse_policy}
\end{algorithm}

\begin{figure*}[t]
\centering
\includegraphics[width=.30\linewidth]{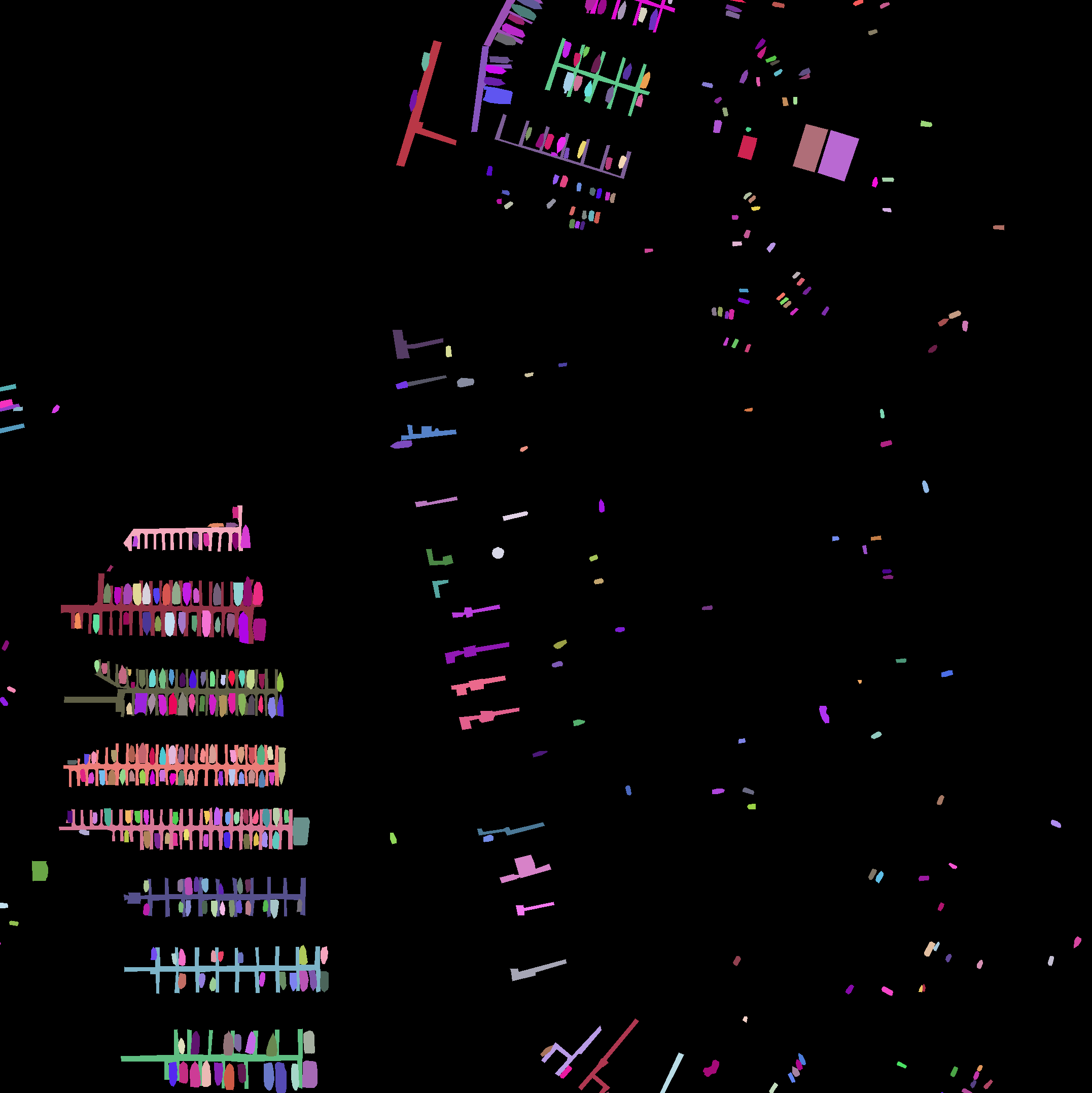}
\includegraphics[width=.30\linewidth]{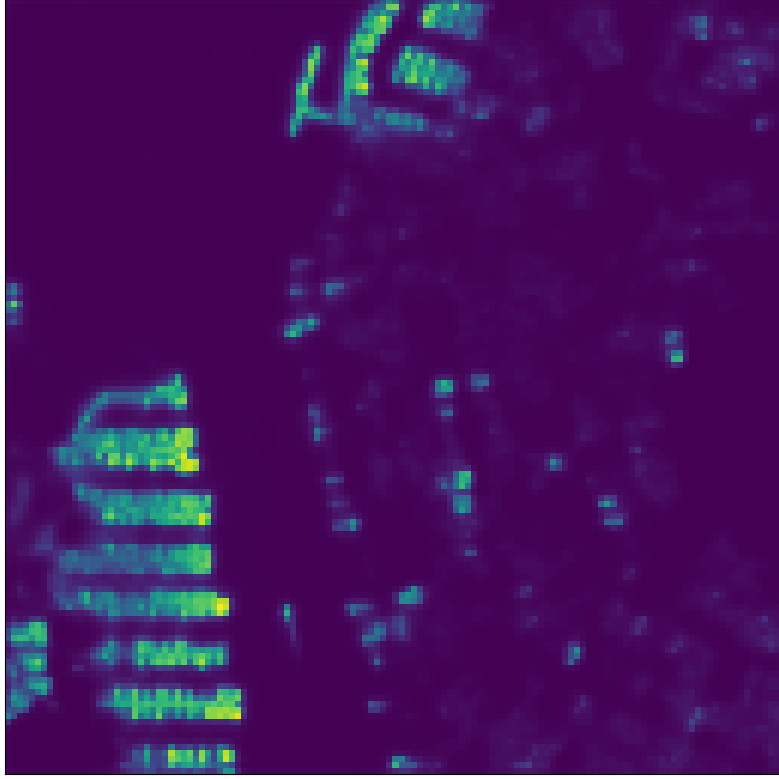}
\includegraphics[width=.30\linewidth]{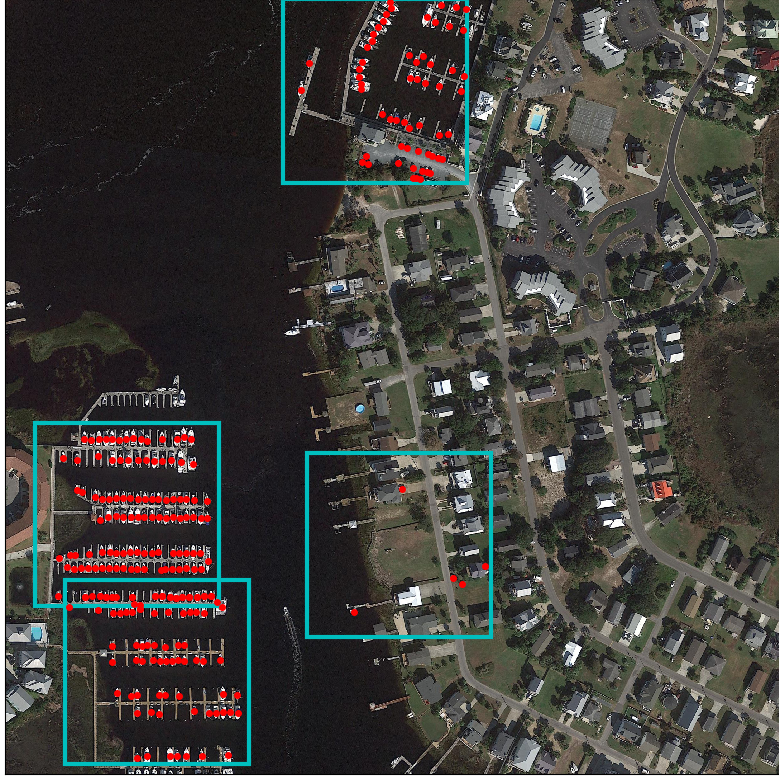}
\caption{(left) Ground truth instance annotation for all DOTA classes. (middle) Low-resolution gist objectness map produced by the U-Net trained on DOTA6. (right) Original image with glimpse (cyan boxes) selections and detections (red dots).  For a fixed budget of four HR windows, this example illustrates the potential for objectness-guided search to search the most likely occupied regions of interest first. Note that images are for visualization purposes only and are not shown to scale.}
\label{fig:objectness_eval}
\end{figure*}

\subsection{Efficient Open-Set Object Detection and Target-Guided Search}

We now turn to the task of target-based, open-set search. In this setting the goal is to search the image for tiles (sub-images) that support the presence of a desired target object, while relying on only a single example image of that target (which we denote here as $o$). During both training and inference, the detector model is presented with a target image from which it must implicitly infer a previously unseen target class and a search image in which it must localize instances of the target object and produce a set of tile locations containing that target.

Consider a set of candidate search locations obtained by tiling the original image into a set of sub-image tiles and denote their coordinates as $(i,j)$. To achieve the goal of optimal open-set search, we use as criteria the likelihood  $l_o(i,j)$ that the specific target object be found at that location. We estimate this likelihood by comparing the network embedding representations $f$ of the exemplar object image $o$ and the tile $T(i,j)$:
 
\begin{equation}
l_o(i,j) = \cos(f(T(i,j)), f(o)).
\end{equation}

\noindent We use the second-to-last layer of ResNet50 as the embedding function $f(.)$ in computing this {\it deep similarity} $l_o(i,j)$.

We further consider a refined approach that uses a  maximum a posteriori criteria, wherein we interpret the objectness probability $\pi_{i,j}$ (in Eq.~(\ref{eq:obj_loss})), integrated over a specific tile (at coordinates $(i,j)$), as a prior that any object of interest be present in that tile. The resulting a posteriori probability $p_o(i,j)$ that the target object is present in the tile is then:

\begin{equation}
p_o(i,j) = l_o(i,j) \pi_{i,j}.
\end{equation}

We use these above metrics to obtain a rank ordering by deep similarity of tiles to visit. Our search policy is chosen simply to visit tiles in decreasing order for either $l_o(i,j)$ or  $p_o(i,j)$.  We call global searches ordered by likelihood $l_o(i,j)$ G-ML-MSTR (``global search / searching for most similar to reference'') and global searchers ordered by a-posteriori probability $p_o(i,j)$ G-MAP-MSTR.

\section{Experiments}
\subsection{Data}
To test our objectness-based approach, we run an evaluation on the DOTA/iSAID overhead dataset~\cite{Xia_2018_CVPR,waqas2019isaid} which consists of 2806 overhead images ranging in size from 800x800 to 4000x4000. In order to test the ability of our method to find objects that are small relative to the size of the image, we narrow our evaluation to six object classes: \textit{small vehicle, large vehicle, plane, helicopter, ship, and storage tank} (henceforth referred to as DOTA6).  

We train all core models on the DOTA training split and evaluate on the images of the validation split.  Of the 458 images in the validation split, we test on only those containing at least one instance of the objects in DOTA6. 

\begin{figure*}[h]
\centering
\includegraphics[width=.32\linewidth]{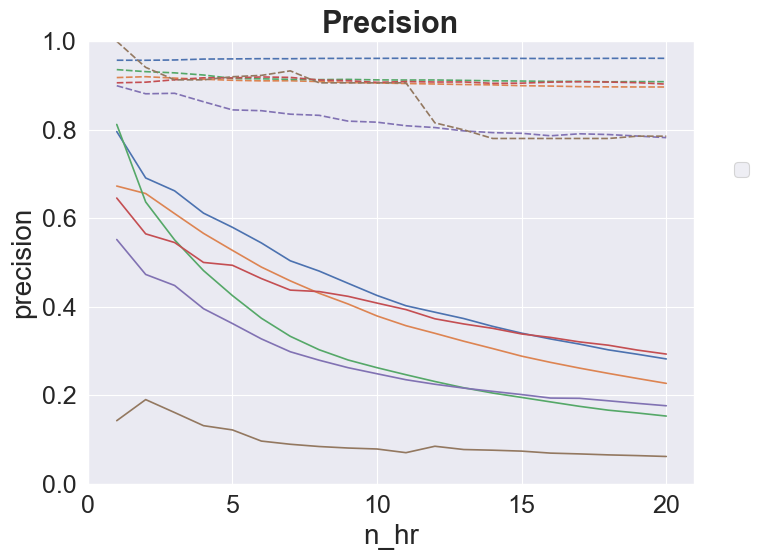}
\includegraphics[width=.32\linewidth]{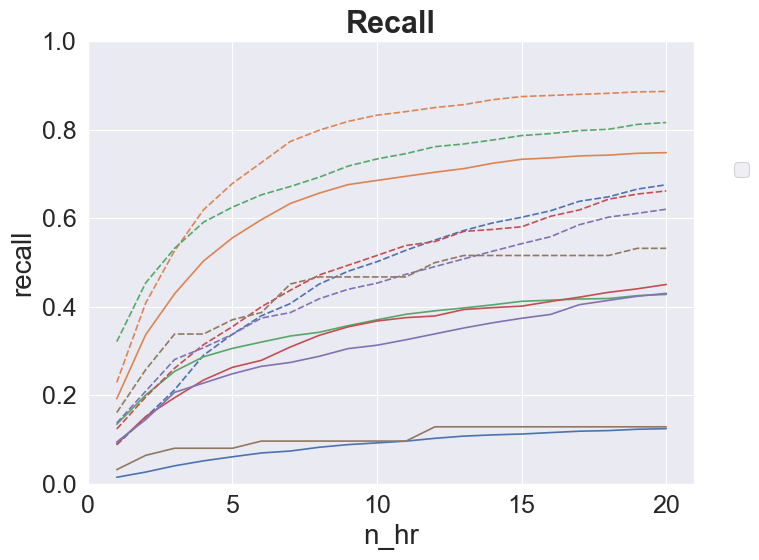}
\includegraphics[width=.33\linewidth]{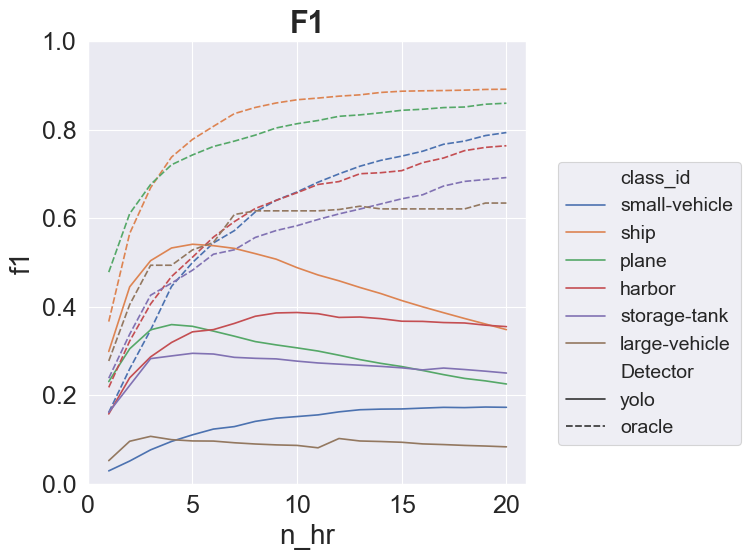}
\caption{Precision, recall, and F1 score by object class in DOTA6 vs. the number of HR glimpses evaluated. Results are separated by detection method (YOLOv3 vs. best-case oracle). The best case is that each metric reaches 1.0 in the fewest number of glimpses (denoted as $n_{hr}$).}
\label{fig:objectness_eval}
\end{figure*}

\begin{figure*}[h]
\centering
\includegraphics[width=.32\linewidth]{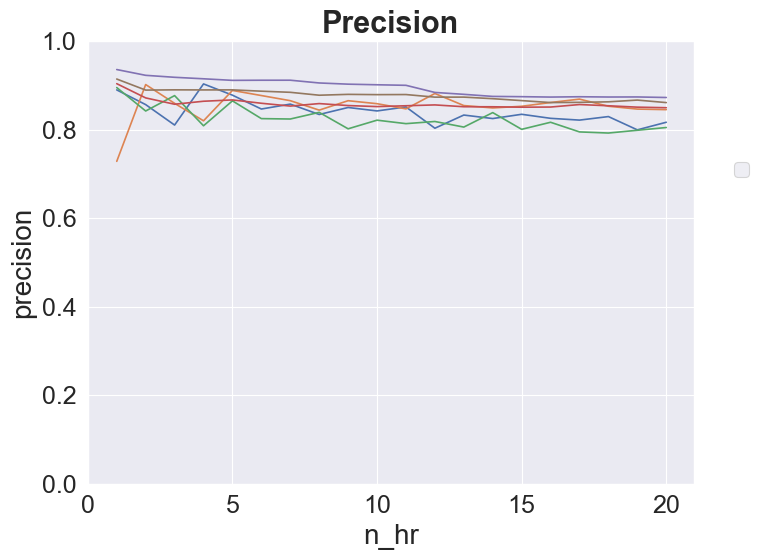}
\includegraphics[width=.32\linewidth]{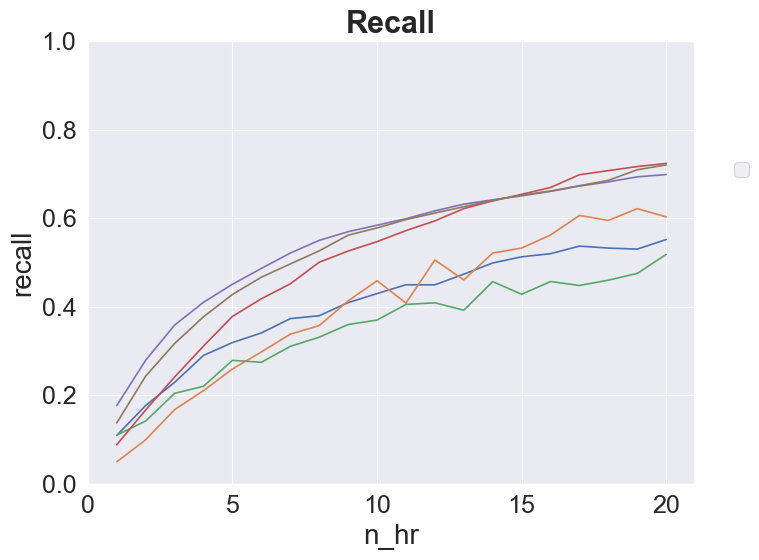}
\includegraphics[width=.33\linewidth]{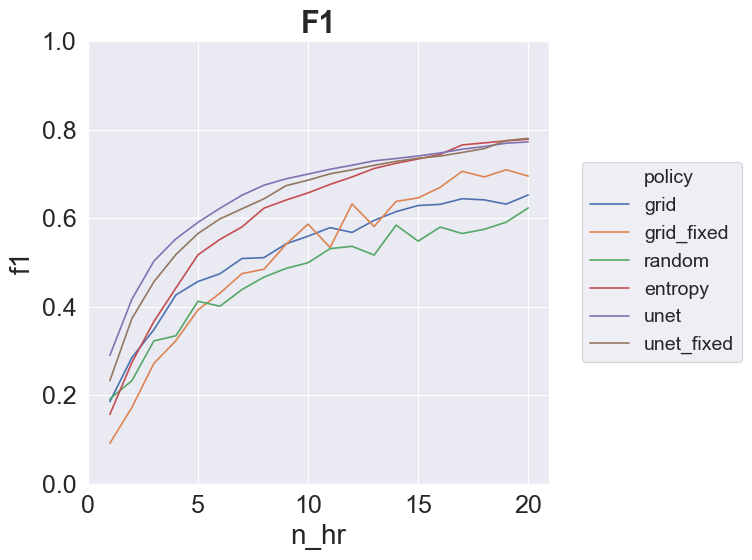}
\caption{Precision, recall, and F1 score averaged over object classes in DOTA6 vs. the number of HR glimpses evaluated. Results are presented using the oracle detector.The best case is that each metric reaches 1.0 in the fewest number of glimpses (denoted as $n_{hr}$).}
\label{fig:oracle_eval}
\end{figure*}

\subsubsection{Baseline Object Detectors}
\label{sec:detectors}
In this work, we use a trained YOLOv3 detector operating on 512x512 tiles.  Because we do not make any algorithmic modifications to the HR-detector architecture itself, the quality of the detection results hinges on our algorithm's ability to select HR glimpses to evaluate.  As such, our baselines were established to measure the quality of the glimpse selection.  We implemented several policies to sample glimpses either from a fixed grid or flexibly (allowing possible overlap of glimpses). Baselines are described in Table~\ref{tab:baselines}.

\subsection{Closed-set Objectness-guided Detection}
\subsubsection{Objectness U-Net}

For estimating objectness, we use a U-Net architecture based on four downsampling and four upsampling blocks, where each block down-/up-samples the input feature map by a factor of two and then passes the result through two blocks of \{convolution, batch normalization, ReLU\}. Inputs to the U-Net model also pass through a similar block before undergoing downsampling. At each downsampling block the number of channels is doubled, starting at 64 and ending at 512. Upsampling blocks concatenate the upsampled feature maps with the maps from the skip connection (achieving twice the original number of channels at the corresponding level in the downsampling path) and produce feature maps with 512 channels at the lowest point of the U-Net down to 64 channels at the highest point. Our U-Net accepts inputs of dimensions 128x128x3 and returns a 1-channel objectness prediction of dimension 128x128. 

Segmentation masks for the DOTA6 labels were converted to a binary map and our U-Net was trained via backpropagation using ADAM~\cite{kingma2014adam} optimization with a standard Binary Cross Entropy loss (Eq.~(\ref{eq:obj_loss})) between the objectness prediction and ground-truth mask.

\begin{table}[h]
\scriptsize
    \begin{tabular}{p{1.0cm}|p{6.8cm}}
        \hline
         \normalsize{\textbf{Policy}} & \normalsize{\textbf{Description}} \\
         \hline 
         grid & Glimpses spaced uniformly with possible overlap for large $n_{\text{glimpse}}$ \\
         \hline 
         grid\_fixed & Non-overlapping tiles ordered randomly \\
         \hline 
         random & Overlapping glimpses whose coordinates are sampled uniformly \\
         \hline 
         entropy & Glimpses sampled according to entropy (highest to lowest) \\
         \hline 
         \textbf{unet} & Objectness-guided search according to Algorithm~\ref{alg:glimpse_policy} \\
         \hline 
         \textbf{unet\_fixed} & Objectness-guided (highest to lowest) search of non-overlapping tiles \\
         \hline 
    \end{tabular}
    \caption{Glimpse selection policies}
    \label{tab:baselines}
\end{table}
\vspace{-.4cm}

\subsubsection{Closed-Set Detection Results}
We evaluate our method using two approaches: (1) a trained, state-of-the-art object detection model (see~\ref{sec:detectors}) and (2) an oracle detector that correctly detects anything within a sampled high-resolution glimpse.  In so doing we decouple the effects of the glimpsing and object detection strategies.

Additionally, in order to address computational efficiency, we evaluate detection performance considering cases with a fixed HR glimpse budget.  Results are shown in Figure~\ref{fig:objectness_eval}. We determine true/false detections following the standard practice of using an Intersection-over-Union (IoU) threshold of 0.5 between predicted and true bounding boxes.

Figure~\ref{fig:oracle_eval} shows performance averaged across all DOTA6 classes.  Note that while the oracle detector (discussed later) represents best case performance, it is still unable to achieve perfect precision since there are cases where sampled glimpses contain partial objects that fail to meet the IoU detection threshold.

\subsection{Open-Set Detection and Target Guided Search}

We again use DOTA6 to evaluate our open-set search strategies against the following baselines: a naive global sliding window and local searches in predefined regions maximizing similarity to the target tile, similarity to the initial location,  and similarity to the current tile.  The local search is motivated by the observation that objects of interest tend to cluster in specific regions (e.g. cars clustered in a parking lot).

\begin{figure}[h!]
\centering
\includegraphics[width=6.6cm]{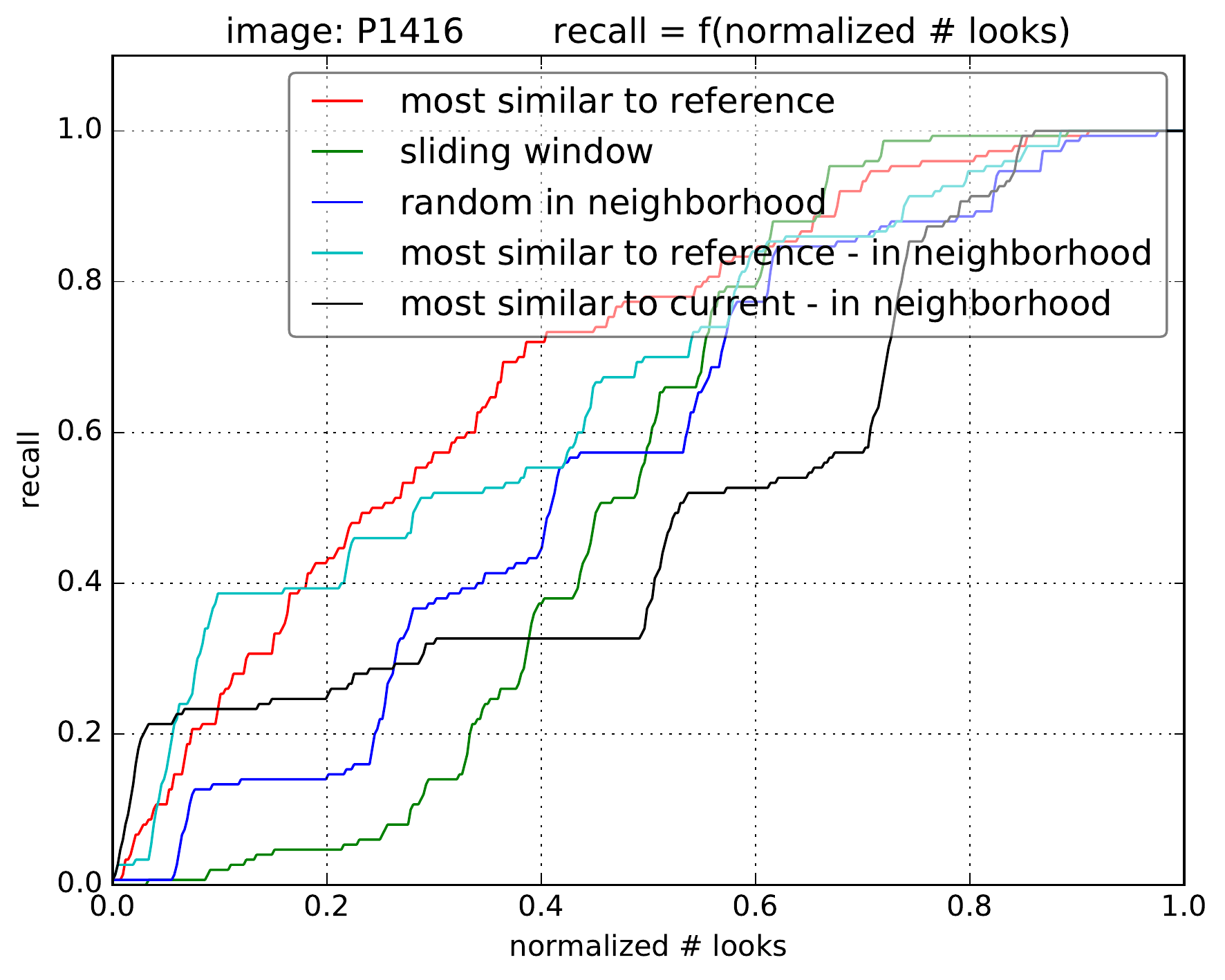}\hspace{-0.4cm}
\hspace{1cm}
\includegraphics[width=6cm]{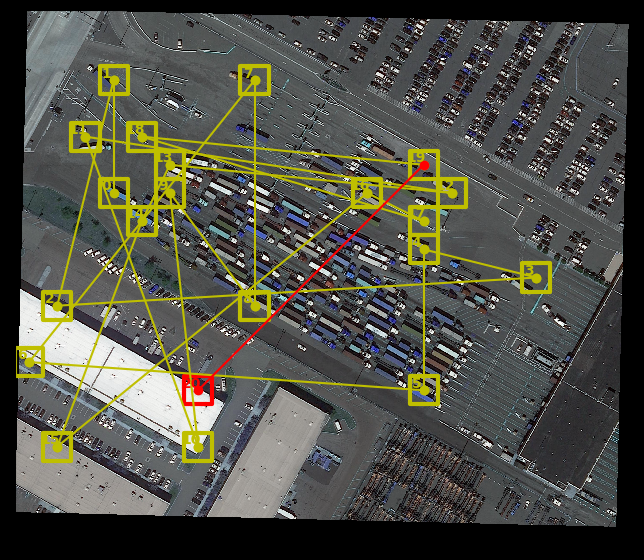}\hspace{-0.4cm}
\caption{(top) performance comparison of MSTR-ML with other baseline approaches: recall vs normalized number of looks. (bottom) illustration of the search trajectory using MSTR-ML.}
\label{fig:image1}
\end{figure}

\begin{figure}[t]
\centering
\includegraphics[width=6.6cm]{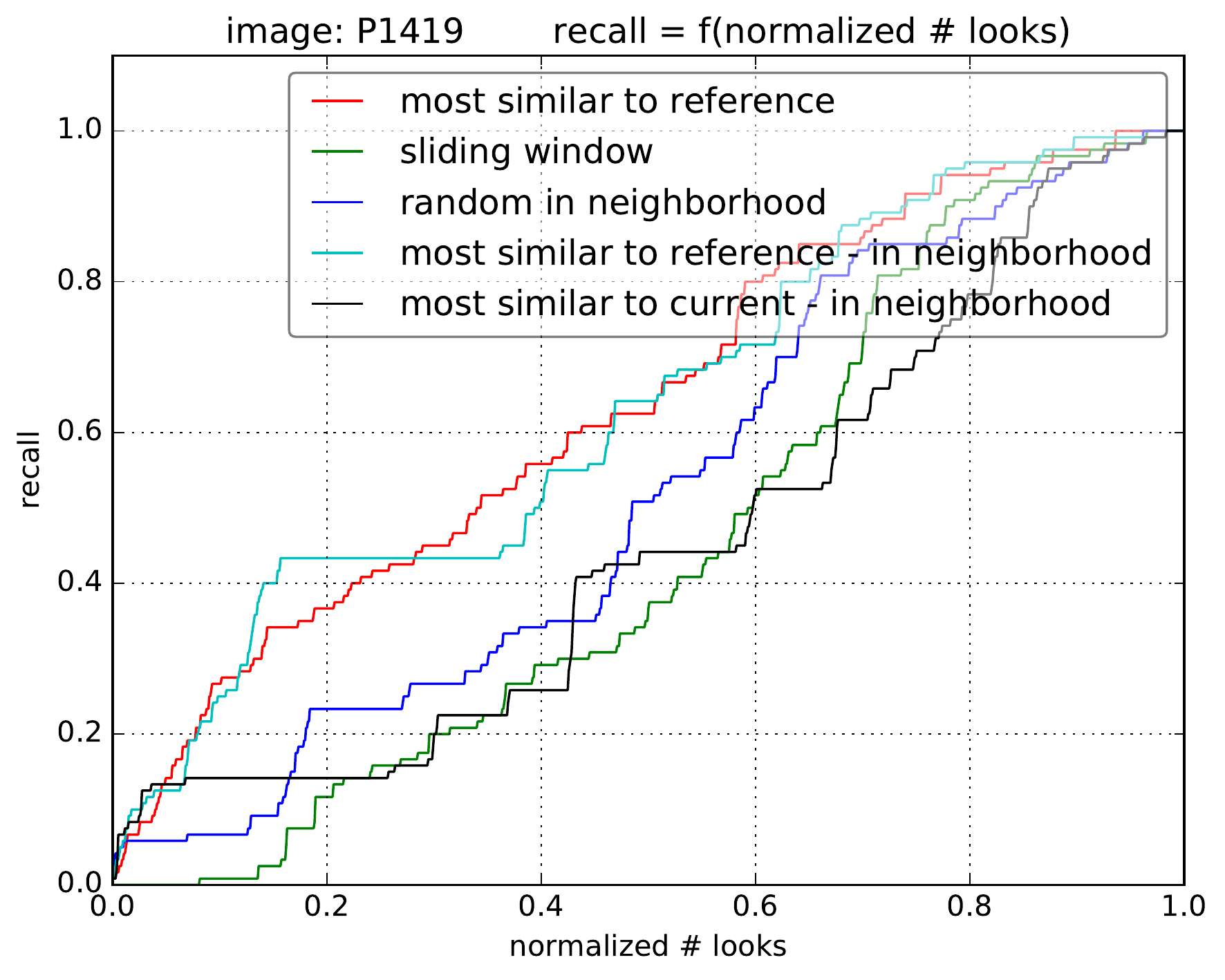}\hspace{-0.4cm}
\hspace{1cm}
\includegraphics[width=5.6cm]{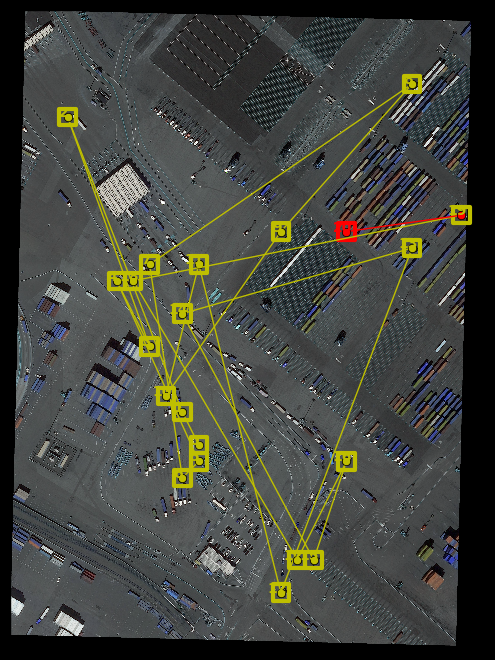}\hspace{-0.4cm}
\caption{(top) performance comparison of MSTR-ML with other baseline approaches: recall vs normalized number of looks. (bottom) illustration of the search trajectory using MSTR-ML.}
\label{fig:image4}
\end{figure}

\begin{figure}[t]
\centering
\includegraphics[width=6.6cm]{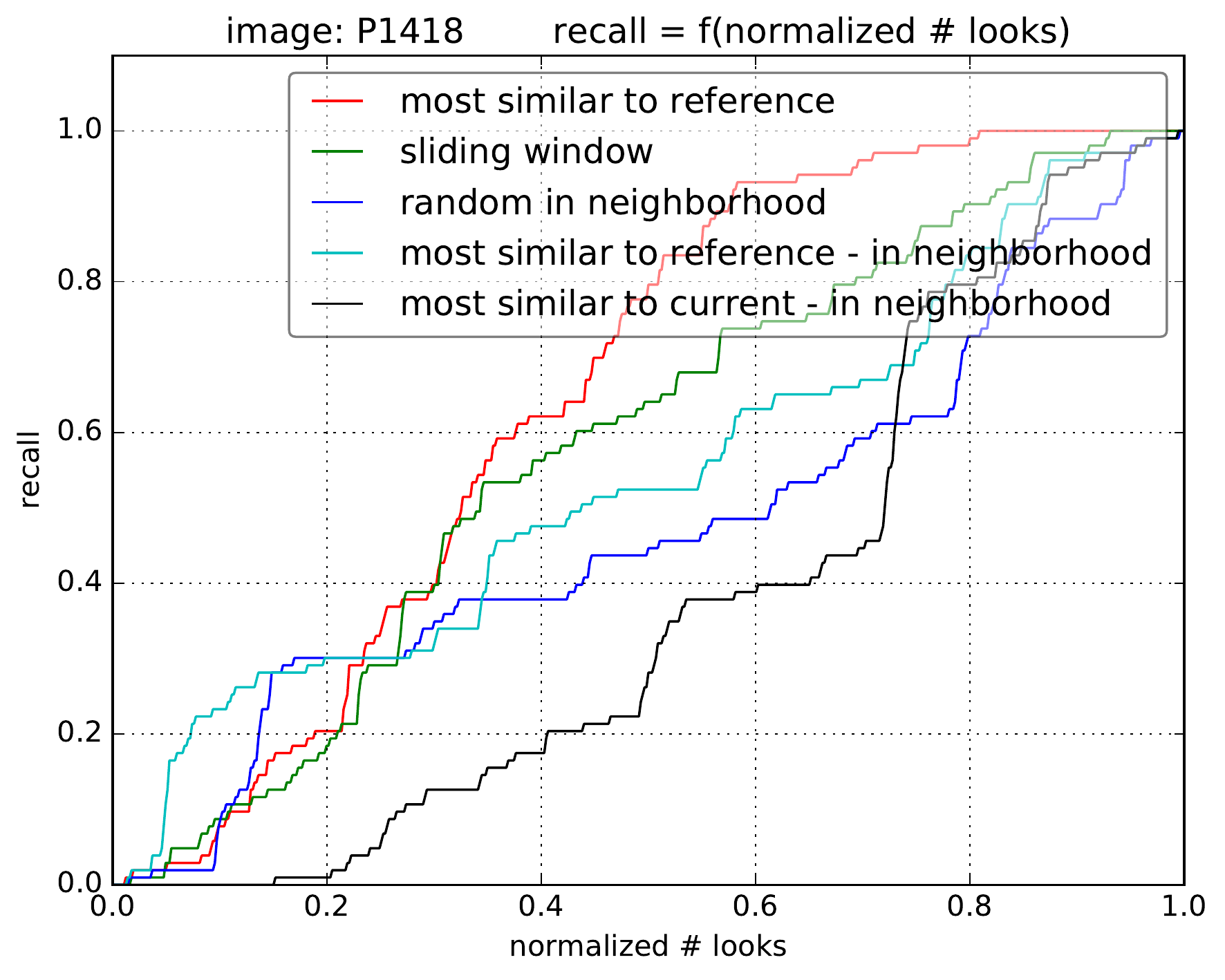}\hspace{-0.4cm}
\hspace{1cm}
\includegraphics[width=5.6cm]{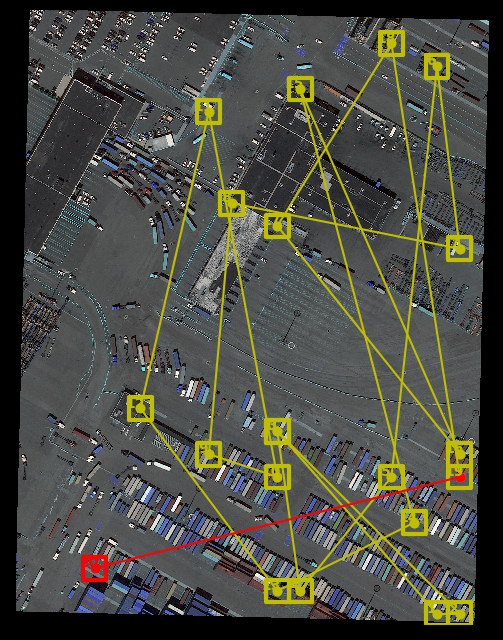}\hspace{-0.4cm}
\caption{(top) performance comparison of MSTR-ML with other baseline approaches: recall vs normalized number of looks. (bottom) illustration of the search trajectory using MSTR-ML.}
\label{fig:image3}
\end{figure}

The sequences of search trajectories that these approaches generate  are compared -- for illustrative purposes -- in the lower panels of Figures \ref{fig:image1}, \ref{fig:image4}, and \ref{fig:image3}. In the top panes of these figures, we plot the recall value as a function of the number of search steps. Here we define recall in terms of the ground truth locations and assume that the algorithm recognizes an object when it lands upon it (i.e., oracle detection).

\begin{figure}[h!]
\centering
\includegraphics[width=\linewidth]{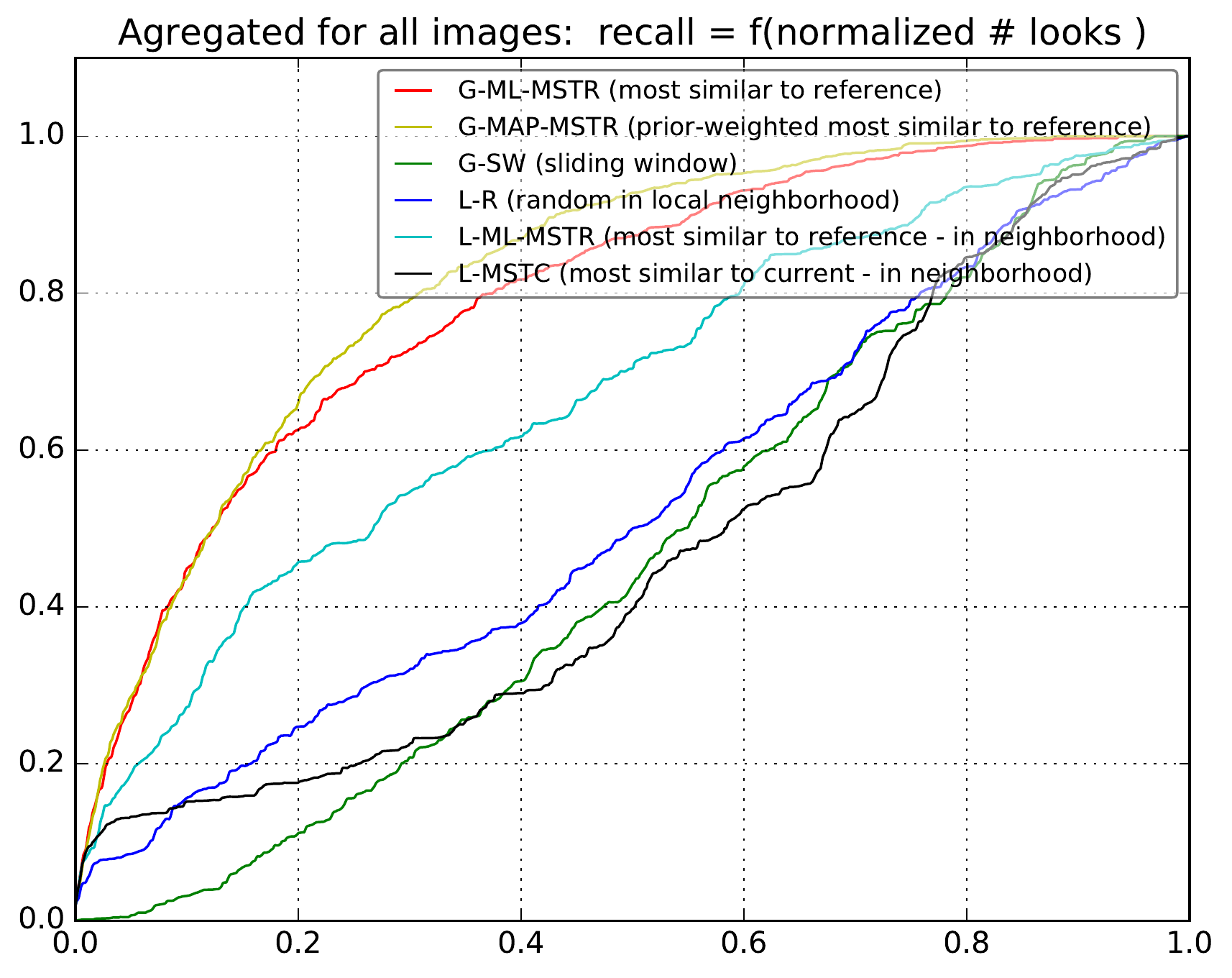}\hspace{-0.4cm}
\caption{Performance (recall) comparison aggregated over all images for  all proposed methods against all baselines described. The y-axis is recall and the x-axis is the normalized number of looks (where 1 indicates that all tiles in the image have been searched and explored.)}
\label{fig:image5}
\end{figure}

An overall comparison of the different strategies over a set of N=11 images is provided in Figure \ref{fig:image5}.  It can be seen that the best performance is achieved by the global ML and MAP-guided approaches, with a clear benefit being observed when the prior $\pi(i,j)$ in used in the selection of the next tile.

\section{Discussion}
Several observations can be made about our approaches and results. 

In the closed-set paradigm, we find that conventional object detectors, such as YOLOv3, exhibit poor performance in the overhead domain.  This is likely due to the differences in object scale and density in overhead imagery compared to the ground-level, natural images for which they were developed.  The situation is further complicated by the inherent variability of the DOTA dataset, which was gathered from a variety of sensors and under a variety of imaging conditions.  DOTA therefore contains significant variation in both image resolution and Ground Sample Distance (GSD).

In both the closed-set object detection and open-set target-guided search tasks, our objectness maps provide a strong prior for improving the efficiency of the search.  The objectness prior improves recall for closed-set detection, particularly in the first few glimpses.  This attribute is important given typical computational budgets and because of the absence of other detection methods for vHR overhead imagery.  In open-set target-guided search, we see that global searches using similarity to the reference representation solidly outperform the baselines.  A noticeable gain is also seen when including the objectiveness prior in that approach. 

While we made considerable improvements over baseline methods, closed-set object detection and open-set search of overhead imagery remain areas ripe for future research. Areas of future work may include extensions of the current work in both the closed- and open-set settings by using objectness along with deep reinforcement learning~\cite{Uzkent2020EfficientOD} and, when carrying out open-set search in the context of  applications in autonomy and robotics applications, the use of Bayesian filtering~\cite{banerjee2010efficient} to improve smoothness of search trajectories.

\section{Conclusion}

This study examines both closed-set object detection and open-set target search. It proposes a method for predicting pixel-level objectness from low resolution gist images, which is then used to choose high-resolution regions for object detection and 
 in a Bayesian approach for open-set visual target search.  The objectness-guided approach is seen to have benefits for improving the efficiency of vHR image processing.  This is true both for selecting HR glimpses and for incorporation as a prior in open set target search.  Both approaches are shown to improve performance when compared to baseline methods.

{\small
\bibliographystyle{ieee_fullname}
\bibliography{egbib}

\begin{thebibliography}{10}\itemsep=-1pt

\bibitem{banerjee2010efficient}
Amit Banerjee and Philippe Burlina.
\newblock Efficient particle filtering via sparse kernel density estimation.
\newblock {\em IEEE Transactions on Image Processing}, 19(9):2480--2490, 2010.

\bibitem{bochkovskiy_yolov4_2020}
Alexey Bochkovskiy, Chien-Yao Wang, and Hong-Yuan~Mark Liao.
\newblock {YOLOv4}: {Optimal} {Speed} and {Accuracy} of {Object} {Detection}.
\newblock {\em arXiv:2004.10934 [cs, eess]}, Apr. 2020.

\bibitem{burlina2011automatic}
Philippe Burlina, David~E Freund, B{\'e}n{\'e}dicte Dupas, and Neil Bressler.
\newblock Automatic screening of age-related macular degeneration and retinal
  abnormalities.
\newblock In {\em 2011 Annual International Conference of the IEEE Engineering
  in Medicine and Biology Society}, pages 3962--3966. IEEE, 2011.

\bibitem{burlina2021addressing}
Philippe Burlina, Neil Joshi, William Paul, Katia~D Pacheco, and Neil~M
  Bressler.
\newblock Addressing artificial intelligence bias in retinal diagnostics.
\newblock {\em Translational Vision Science \& Technology}, 10(2):13--13, 2021.

\bibitem{burlina2020low}
Philippe Burlina, William Paul, Philip Mathew, Neil Joshi, Katia~D Pacheco, and
  Neil~M Bressler.
\newblock Low-shot deep learning of diabetic retinopathy with potential
  applications to address artificial intelligence bias in retinal diagnostics
  and rare ophthalmic diseases.
\newblock {\em JAMA ophthalmology}, 138(10):1070--1077, 2020.

\bibitem{burlina2019assessment}
Philippe~M Burlina, Neil Joshi, Katia~D Pacheco, TY~Alvin Liu, and Neil~M
  Bressler.
\newblock Assessment of deep generative models for high-resolution synthetic
  retinal image generation of age-related macular degeneration.
\newblock {\em JAMA ophthalmology}, 137(3):258--264, 2019.

\bibitem{carlini2017adversarial}
Nicholas Carlini and David Wagner.
\newblock Adversarial examples are not easily detected: Bypassing ten detection
  methods.
\newblock In {\em Proceedings of the 10th ACM Workshop on Artificial
  Intelligence and Security}, pages 3--14, 2017.

\bibitem{imagenet_cvpr09}
J. Deng, W. Dong, R. Socher, L.-J. Li, K. Li, and L. Fei-Fei.
\newblock Imagenet: A large-scale hierarchical image database.
\newblock In {\em CVPR09}, 2009.

\bibitem{vanetten2018}
Adam~Van Etten.
\newblock You only look twice: Rapid multi-scale object detection in satellite
  imagery.
\newblock {\em CoRR}, abs/1805.09512, 2018.

\bibitem{geng2020recent}
Chuanxing Geng, Sheng-jun Huang, and Songcan Chen.
\newblock Recent advances in open set recognition: A survey.
\newblock {\em IEEE Transactions on Pattern Analysis and Machine Intelligence},
  2020.

\bibitem{fastrcnn}
R. {Girshick}.
\newblock Fast r-cnn.
\newblock In {\em 2015 IEEE International Conference on Computer Vision
  (ICCV)}, pages 1440--1448, 2015.

\bibitem{girshick_rich_2014}
Ross Girshick, Jeff Donahue, Trevor Darrell, and Jitendra Malik.
\newblock Rich feature hierarchies for accurate object detection and semantic
  segmentation.
\newblock {\em arXiv:1311.2524 [cs]}, 2014.
\newblock arXiv: 1311.2524.

\bibitem{he2016deep}
Kaiming He, Xiangyu Zhang, Shaoqing Ren, and Jian Sun.
\newblock Deep residual learning for image recognition.
\newblock In {\em Proceedings of the IEEE conference on computer vision and
  pattern recognition}, pages 770--778, 2016.

\bibitem{kingma2014adam}
Diederik~P Kingma and Jimmy Ba.
\newblock Adam: A method for stochastic optimization.
\newblock {\em arXiv preprint arXiv:1412.6980}, 2014.

\bibitem{krizhevsky2017imagenet}
Alex Krizhevsky, Ilya Sutskever, and Geoffrey~E Hinton.
\newblock Imagenet classification with deep convolutional neural networks.
\newblock {\em Communications of the ACM}, 60(6):84--90, 2017.

\bibitem{Liao_2020}
W. Liao, X. Chen, J. Yang, S. Roth, M. Goesele, M.~Y. Yang, and B. Rosenhahn.
\newblock Lr-cnn: Local-aware region cnn for vehicle detection in aerial
  imagery.
\newblock {\em ISPRS Annals of Photogrammetry, Remote Sensing and Spatial
  Information Sciences}, V-2-2020:381–388, Aug 2020.

\bibitem{pekala2019deep}
Mike Pekala, Neil Joshi, TY~Alvin Liu, Neil~M Bressler, D~Cabrera DeBuc, and
  Philippe Burlina.
\newblock Deep learning based retinal oct segmentation.
\newblock {\em Computers in Biology and Medicine}, 114:103445, 2019.

\bibitem{rainey2012}
Katie Rainey, Shibin Parameswaran, Josh Harguess, and John Stastny.
\newblock {Vessel classification in overhead satellite imagery using learned
  dictionaries}.
\newblock In Andrew~G. Tescher, editor, {\em Applications of Digital Image
  Processing XXXV}, volume 8499, pages 741 -- 752. International Society for
  Optics and Photonics, SPIE, 2012.

\bibitem{ravi2016optimization}
Sachin Ravi and Hugo Larochelle.
\newblock Optimization as a model for few-shot learning.
\newblock 2016.

\bibitem{Redmon2015YouOL}
Joseph Redmon, Santosh~Kumar Divvala, Ross~B. Girshick, and Ali Farhadi.
\newblock You only look once: Unified, real-time object detection.
\newblock {\em 2016 IEEE Conference on Computer Vision and Pattern Recognition
  (CVPR)}, pages 779--788, 2015.

\bibitem{redmon2018yolov3}
Joseph Redmon and Ali Farhadi.
\newblock Yolov3: An incremental improvement, 2018.
\newblock cite arxiv:1804.02767Comment: Tech Report.

\bibitem{Ren2015FasterRT}
Shaoqing Ren, Kaiming He, Ross~B. Girshick, and Jian Sun.
\newblock Faster r-cnn: Towards real-time object detection with region proposal
  networks.
\newblock {\em IEEE Transactions on Pattern Analysis and Machine Intelligence},
  39:1137--1149, 2015.

\bibitem{ronneberger2015u}
Olaf Ronneberger, Philipp Fischer, and Thomas Brox.
\newblock U-net: Convolutional networks for biomedical image segmentation.
\newblock In {\em International Conference on Medical image computing and
  computer-assisted intervention}, pages 234--241. Springer, 2015.

\bibitem{sakla2017}
W. {Sakla}, G. {Konjevod}, and T.~N. {Mundhenk}.
\newblock Deep multi-modal vehicle detection in aerial isr imagery.
\newblock In {\em 2017 IEEE Winter Conference on Applications of Computer
  Vision (WACV)}, pages 916--923, 2017.

\bibitem{scheirer2012toward}
Walter~J Scheirer, Anderson de Rezende~Rocha, Archana Sapkota, and Terrance~E
  Boult.
\newblock Toward open set recognition.
\newblock {\em IEEE transactions on pattern analysis and machine intelligence},
  35(7):1757--1772, 2012.

\bibitem{Shermeyer_2019_CVPR_Workshops}
Jacob Shermeyer and Adam Van~Etten.
\newblock The effects of super-resolution on object detection performance in
  satellite imagery.
\newblock In {\em Proceedings of the IEEE/CVF Conference on Computer Vision and
  Pattern Recognition (CVPR) Workshops}, June 2019.

\bibitem{shokri2017membership}
Reza Shokri, Marco Stronati, Congzheng Song, and Vitaly Shmatikov.
\newblock Membership inference attacks against machine learning models.
\newblock In {\em 2017 IEEE Symposium on Security and Privacy (SP)}, pages
  3--18. IEEE, 2017.

\bibitem{takemura2015synaptic}
Shin-ya Takemura, C~Shan Xu, Zhiyuan Lu, Patricia~K Rivlin, Toufiq Parag,
  Donald~J Olbris, Stephen Plaza, Ting Zhao, William~T Katz, Lowell Umayam,
  et~al.
\newblock Synaptic circuits and their variations within different columns in
  the visual system of drosophila.
\newblock {\em Proceedings of the National Academy of Sciences},
  112(44):13711--13716, 2015.

\bibitem{Tan_2020_CVPR}
Mingxing Tan, Ruoming Pang, and Quoc~V. Le.
\newblock Efficientdet: Scalable and efficient object detection.
\newblock In {\em The IEEE/CVF Conference on Computer Vision and Pattern
  Recognition (CVPR)}, June 2020.

\bibitem{tang2017vehicle}
Tianyu Tang, Shilin Zhou, Zhipeng Deng, Huanxin Zou, and Lin Lei.
\newblock Vehicle detection in aerial images based on region convolutional
  neural networks and hard negative example mining.
\newblock {\em Sensors}, 17(2):336, 2017.

\bibitem{Uzkent2020EfficientOD}
Burak Uzkent, Christopher Yeh, and Stefano Ermon.
\newblock Efficient object detection in large images using deep reinforcement
  learning.
\newblock {\em 2020 IEEE Winter Conference on Applications of Computer Vision
  (WACV)}, pages 1813--1822, 2020.

\bibitem{vanetten2019}
A. {Van Etten}.
\newblock Satellite imagery multiscale rapid detection with windowed networks.
\newblock In {\em 2019 IEEE Winter Conference on Applications of Computer
  Vision (WACV)}, pages 735--743, 2019.

\bibitem{waqas2019isaid}
Syed Waqas~Zamir, Aditya Arora, Akshita Gupta, Salman Khan, Guolei Sun, Fahad
  Shahbaz~Khan, Fan Zhu, Ling Shao, Gui-Song Xia, and Xiang Bai.
\newblock isaid: A large-scale dataset for instance segmentation in aerial
  images.
\newblock In {\em Proceedings of the IEEE Conference on Computer Vision and
  Pattern Recognition Workshops}, pages 28--37, 2019.

\bibitem{Xia_2018_CVPR}
Gui-Song Xia, Xiang Bai, Jian Ding, Zhen Zhu, Serge Belongie, Jiebo Luo, Mihai
  Datcu, Marcello Pelillo, and Liangpei Zhang.
\newblock Dota: A large-scale dataset for object detection in aerial images.
\newblock In {\em The IEEE Conference on Computer Vision and Pattern
  Recognition (CVPR)}, June 2018.

\end{thebibliography}
}

\end{document}